\documentclass[11pt]{article}

\usepackage{acl}

\usepackage{times}
\usepackage{latexsym}
\usepackage[T1]{fontenc}
\usepackage[utf8]{inputenc}
\usepackage{microtype}
\usepackage{inconsolata}

\usepackage{graphicx}
\usepackage{amsmath,amssymb,amsfonts}
\usepackage{booktabs}
\usepackage{multirow}
\usepackage{array}
\usepackage{makecell}
\usepackage{xcolor}
\usepackage{dblfloatfix}
\usepackage{placeins}
\title{Locality Matters for Training-Free Audio Token Compression in Audio-Language Models}

\author{
Jiale Luo\textsuperscript{1},
Xiaoyu Liang\textsuperscript{1},
Haoji Hu\textsuperscript{1}\thanks{Corresponding author.} \\
\textsuperscript{1}Zhejiang University \\
Correspondence: \texttt{haoji\_hu@zju.edu.cn}
}

\begin{document}
\maketitle

\begin{abstract}
Audio-language models (ALMs) are increasingly used for audio captioning, question answering, and open-ended audio understanding, but their inference cost remains high when audio inputs are represented as long prefix-token sequences. These audio prefixes consume context budget, increase memory usage, and make deployment harder in resource-constrained or latency-sensitive settings. Existing training-free audio-token reduction methods mainly rely on fixed pooling or score-based pruning. Fixed pooling is content-agnostic, while score-based pruning can preserve isolated salient tokens but discard nearby acoustic context. We propose \emph{Local Temporal Bipartite Merging} (LTBM), a training-free encoder-space compression method that merges similar nearby audio tokens under an explicit temporal window constraint. Beyond introducing LTBM, we use a controlled \emph{Global Merge} variant to isolate whether temporal locality itself is a useful inductive bias for audio-token compression. Experiments on AudioCaps, Clotho, and MMAU with Qwen2-Audio show evidence of a task-dependent locality effect: locality-aware merging is more favorable for captioning at several compression settings, especially under stronger compression, while global matching is more competitive for multiple-choice audio understanding. A cross-backbone validation on Audio Flamingo 3 further supports the captioning-side advantage of locality-aware merging under moderate and aggressive compression.
\end{abstract}
\begin{figure*}[t]
    \centering
    \includegraphics[width=0.92\textwidth]{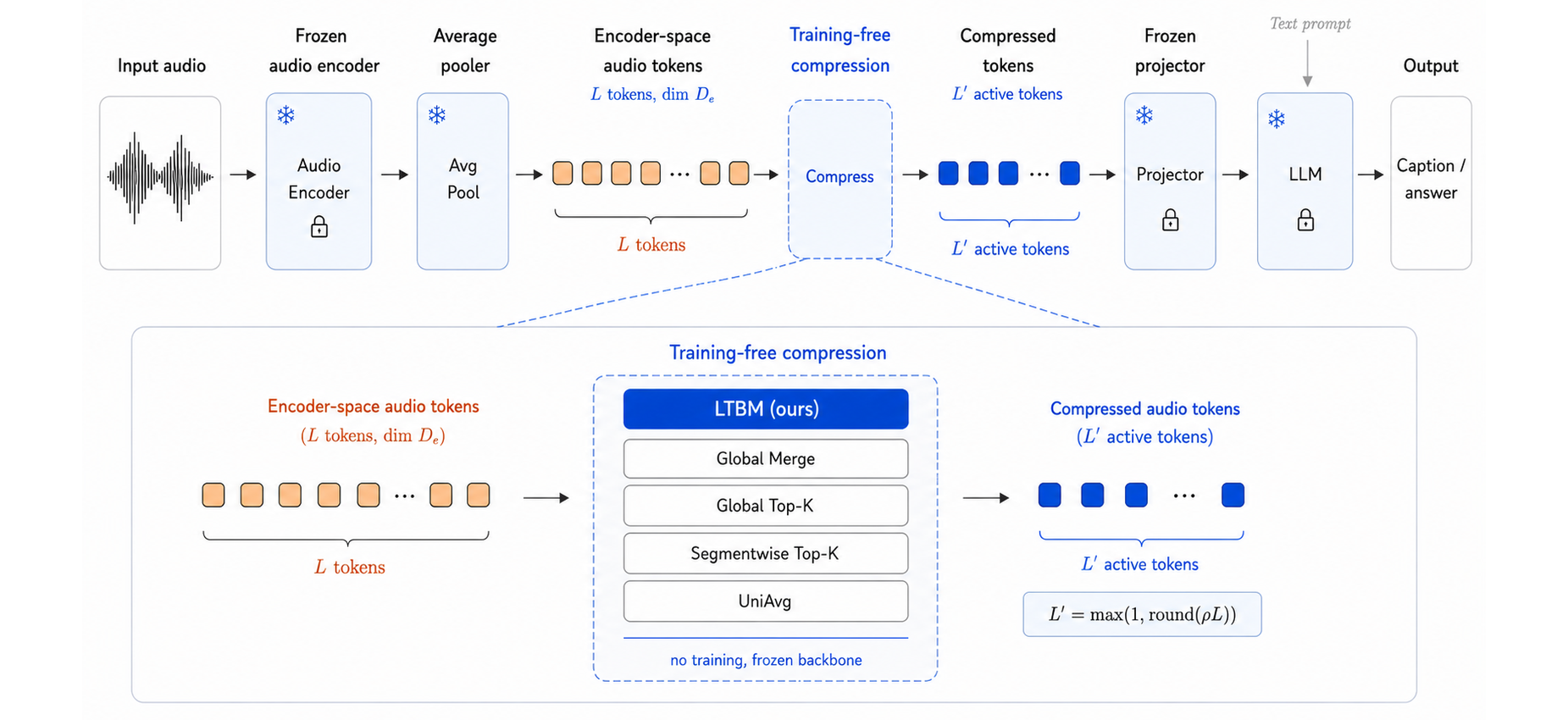}
    \caption{Overview of the training-free audio-token compression pipeline. Compression is applied after the average pooler and before the multimodal projector. The audio encoder, projector, and language model remain frozen, and different compression strategies are compared under the same keep ratio.}
    \label{fig:framework}
\end{figure*}

\section{Introduction}

Audio-language models (ALMs) have become a common interface for audio captioning, question answering, instruction following, and open-ended audio understanding \citep{qwenaudio,qwen2audio,salmonn,audioflamingo,audiopalm,pengi,ltu}. However, their inference cost remains a practical obstacle in resource-constrained and latency-sensitive settings. A key source of this cost is the audio prefix: many ALMs encode an audio clip into a long sequence of continuous tokens, project these tokens into the language-model embedding space, and concatenate them with the textual prompt before autoregressive decoding \citep{qwenaudio,qwen2audio,speechprompt,salmonn}. Compared with a short text prompt, the audio prefix can consume substantially more context budget and significantly increase memory usage and generation cost.

This prefix-based interface is effective and easy to combine with pretrained language models, but its cost grows with audio duration and token density. Even after convolutional subsampling and temporal pooling, an audio clip can still produce hundreds of prefix tokens. Reducing redundant audio tokens without retraining the backbone is therefore an important practical problem for efficient audio-language model inference.

This has motivated token reduction methods for multimodal models, including pooling, pruning, and merging strategies \citep{tome,pumer,llavaprumerge,aim,tokenpacker}. Recent efficient VLM inference studies further explore training-free pruning, dynamic sparsification, and token redundancy reduction \citep{fastervlm,dynamicllava,fitprune,lftr,fastv,tokenpruningmllm,dart}. In audio-language models, recent work has also studied audio-token compression and pruning from both adaptation-based and training-free perspectives \citep{bhati2025atc,gibier2025segmentwise}. We focus on the training-free setting, where the pretrained ALM backbone is kept entirely unchanged.

Existing audio-token reduction methods differ in how they reduce the audio prefix. Fixed pooling methods, such as uniform average pooling, aggregate tokens according to a preset stride and are efficient but content-agnostic \citep{bhati2025atc}. Score-based pruning methods retain tokens with high importance scores, either globally or within temporal segments, and recent ALM pruning work further explores attention-based segmentwise selection \citep{gibier2025segmentwise}. These approaches are adaptive, but pruning performs hard deletion: unselected tokens are removed entirely. For audio, this can be limiting because many acoustic events unfold over short continuous spans rather than as isolated peaks.

This observation motivates token merging as a complementary alternative to hard token selection. Instead of selecting a subset of tokens or averaging fixed intervals, merging can combine similar tokens while preserving aggregated acoustic evidence. However, for audio, the way tokens are merged matters. A globally similar token pair may be far apart in time and correspond to different event occurrences. This makes audio compression different from unordered redundancy removal: similarity should be constrained by temporal order when the goal is to preserve event continuity. Merging such distant tokens can weaken the temporal structure that the language model receives. This raises a more specific guiding question: \emph{does temporal locality matter for training-free audio-token compression?}

To study this question, we propose \emph{Local Temporal Bipartite Merging} (LTBM), a training-free encoder-space compression method that restricts bipartite token matching to a local temporal window. LTBM is content-aware because it merges tokens according to feature similarity, and locality-aware because candidate matches must also be close in time. To isolate the role of locality, we further introduce a controlled \emph{Global Merge} variant that uses the same bipartite merging framework but removes the temporal window constraint. The comparison between LTBM and Global Merge makes locality the central variable rather than an incidental implementation detail.

Our experiments show that locality is useful in a task-dependent manner. On captioning benchmarks, locality-aware merging is more favorable under stronger compression, suggesting that preserving short-range acoustic continuity helps open-ended generation. On MMAU, a multiple-choice audio understanding benchmark, Global Merge is more competitive, suggesting that closed-form decision making may benefit more from globally aggregated evidence. We further evaluate Audio Flamingo 3 as a second backbone and find that the captioning-side advantage of LTBM over Global Merge persists on AudioCaps under moderate and aggressive compression. These results frame locality as a task-dependent design axis for training-free audio-token compression.

Our contributions are as follows:
\begin{itemize}
    \item We propose \textbf{LTBM}, a training-free encoder-space audio-token compression method based on \textbf{local temporal bipartite merging}.
    \item We introduce a controlled \textbf{local-vs-global} comparison through a Global Merge variant that shares the same bipartite merging framework but removes the explicit temporal locality constraint.
    \item With Qwen2-Audio on \textbf{AudioCaps}, \textbf{Clotho}, and \textbf{MMAU}, we identify evidence of a \textbf{task-dependent locality effect}: locality-aware merging is more favorable for captioning at several compression settings, especially under stronger compression, while global matching is more competitive for multiple-choice audio understanding.
    \item We provide a second-backbone validation on \textbf{Audio Flamingo 3}, showing that the captioning-side advantage of locality-aware merging persists under moderate and aggressive compression.
\end{itemize}

\section{Related Work}

\subsection{Audio-Language Models}

Audio-language models connect audio encoders with large language models to support audio captioning, speech and audio understanding, instruction following, and dialogue over acoustic inputs \citep{qwenaudio,qwen2audio,salmonn,audioflamingo,audiopalm,pengi,ltu}. More recent systems extend this paradigm toward long-audio understanding, speech-audio joint reasoning, and open audio-language interaction \citep{qwen25omni,audioflamingo2,audioflamingo3,ltuas,wavllm,gama}. Most of these models rely on a prefix-style interface: audio is encoded into a sequence of continuous tokens, projected into the language-model embedding space, and prepended to the textual prompt. Our work is orthogonal to backbone design. Instead of training a new ALM or modifying its parameters, we study training-free compression of the audio prefix during inference.

\subsection{Efficient Multimodal Inference}

Token reduction has been widely studied for efficient vision and multimodal language models. Prior work reduces visual-token redundancy through pruning, pooling, merging, or dynamic token selection \citep{tome,pumer,llavaprumerge,aim,tokenpacker,fastervlm,dynamicllava,fitprune,lftr,fastv,tokenpruningmllm,dart}. These methods show that many multimodal prefix tokens are redundant for downstream generation, but most of them are designed around image or video tokens, where spatial redundancy is a dominant structure. Audio tokens differ because the signal is temporally ordered and many events unfold over short continuous spans. This motivates studying whether locality should be treated as an explicit inductive bias for audio-prefix compression.

\subsection{Audio-Token Compression and Pruning}

Audio-token reduction in ALMs has recently received increasing attention. Existing methods include fixed audio-token compression and score-based pruning, including segmentwise selection strategies \citep{bhati2025atc,gibier2025segmentwise}. Fixed pooling is efficient and easy to deploy, but its aggregation pattern is independent of acoustic content, while pruning keeps selected tokens and discards the rest. In contrast, LTBM studies token merging: instead of deleting unselected tokens, it aggregates similar nearby representations. Our key distinction is the controlled comparison between local temporal bipartite merging and global bipartite merging, which isolates whether temporal locality itself is useful for training-free audio-prefix compression.

\begin{figure*}[t]
    \centering
    \includegraphics[width=0.92\textwidth]{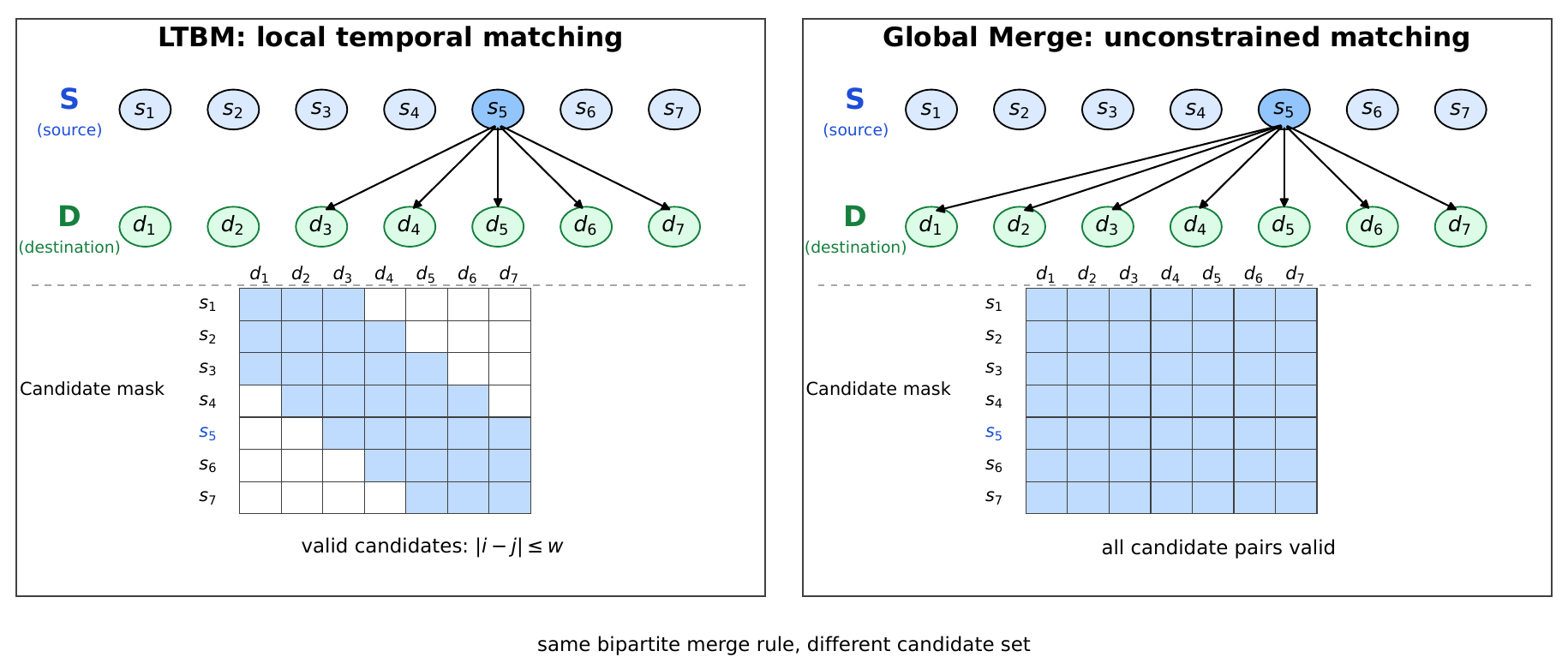}
    \caption{Local versus global bipartite matching. The top diagrams illustrate candidate destinations for one source token, while the bottom masks show valid source--destination candidate pairs. LTBM restricts matching to a local temporal window, whereas Global Merge uses the same bipartite merge rule but allows candidates across the full sequence. The merge step is applied iteratively until the target length $L'=\max(1,\mathrm{round}(\rho L))$ is reached.}
    \label{fig:local_global_vis}
\end{figure*}

\section{Method}

\subsection{Problem Setup}

We consider an ALM composed of an audio encoder, an average pooler, a multimodal projector, and a language model. Given an input audio clip, the encoder and pooler produce a sequence
\begin{equation}
\mathbf{X} = [\mathbf{x}_1,\mathbf{x}_2,\ldots,\mathbf{x}_L], \qquad \mathbf{x}_\ell \in \mathbb{R}^{D_e},
\end{equation}
where $L$ denotes the encoder-space token length and $D_e$ is the feature dimension. Compression is applied after the average pooler and before the multimodal projector, so the method is training-free and does not modify the pretrained backbone weights, as illustrated in Figure~\ref{fig:framework}.

\subsection{Local Temporal Bipartite Merging}

LTBM partitions the sequence by parity into source and destination subsets:
\begin{equation}
\mathcal{S}=\{\mathbf{x}_1,\mathbf{x}_3,\mathbf{x}_5,\ldots\}, \qquad
\mathcal{D}=\{\mathbf{x}_2,\mathbf{x}_4,\mathbf{x}_6,\ldots\}.
\end{equation}
For each source token $\mathbf{s}_i\in\mathcal{S}$ and destination token $\mathbf{d}_j\in\mathcal{D}$, we compute cosine similarity
\begin{equation}
\mathrm{sim}(i,j)=\frac{\mathbf{s}_i^\top \mathbf{d}_j}{\|\mathbf{s}_i\|\,\|\mathbf{d}_j\|}.
\end{equation}
To preserve temporal locality, matching is restricted to a window of size $w$ in the source--destination order:
\begin{equation}
|i-j| \le w,
\end{equation}
where $i$ and $j$ denote the order indices of tokens in $\mathcal{S}$ and $\mathcal{D}$, respectively. Pairs outside the valid temporal window are masked out.

Each source token selects its best destination within the window. Source tokens are then ranked by their best-match similarity, and the highest-ranked sources are merged until the target token length is reached. Let $\mathcal{M}(j)$ denote the set of source tokens merged into destination $j$. The updated destination token is
\begin{equation}
\tilde{\mathbf{d}}_j =
\frac{1}{c_j}\left(\mathbf{d}_j + \sum_{i \in \mathcal{M}(j)} \mathbf{s}_i \right),
\end{equation}
where $c_j = 1 + |\mathcal{M}(j)|$ includes the destination token and all merged source tokens. Unmatched source tokens are retained, and the compressed sequence is reordered to restore temporal order. 
To target a keep ratio $\rho$, the merge step is applied iteratively until it reaches the target length
\begin{equation}
L'=\max(1,\mathrm{round}(\rho L)).
\end{equation}

The central inductive bias of LTBM is explicit temporal locality. Tokens are not merged solely because they are globally similar; they must also be close in time. This distinguishes LTBM from unconstrained content-only matching and makes locality the key variable studied in this paper. Figure~\ref{fig:local_global_vis} illustrates this contrast by comparing local candidate matching in LTBM with unconstrained matching in Global Merge.

\subsection{Global Merge and Other Baselines}

To isolate the role of locality, we define \textbf{Global Merge}, which uses the same bipartite merging framework as LTBM but removes the temporal window constraint. This yields a controlled local-versus-global comparison under otherwise matched compression mechanics.

We further compare against three training-free baselines: \textbf{UniAvg}, \textbf{Global Top-K}, and \textbf{Segmentwise Top-K}. UniAvg is a fixed uniform average-pooling baseline. The two Top-K baselines use encoder-space $L_2$ norm as the importance score: Global Top-K ranks tokens globally, while Segmentwise Top-K ranks tokens within temporal segments. The segmentwise design is inspired by temporal pruning in ALMs \citep{gibier2025segmentwise}, but we do not reproduce their full attention-based method.

The compressed audio representations are injected through a compatibility-preserving interface that keeps the pretrained backbone unchanged and avoids kernel or cache modifications. This allows all methods to be compared under the same multimodal generation pathway.

\section{Experiments}

\subsection{Experimental Details}

Our main experiments use \textbf{Qwen2-Audio-7B-Instruct} \citep{qwen2audio} as the backbone model. We use the official Hugging Face implementation with trusted remote code, automatic device placement, and eager attention, and apply compression after the encoder average pooler and before the multimodal projector. We additionally conduct a cross-backbone validation on \textbf{Audio Flamingo 3}~\citep{audioflamingo3} to examine whether the captioning-side behavior is specific to Qwen2-Audio.

For \textbf{AudioCaps} and \textbf{Clotho}, the prompt is fixed to \emph{``Generate a caption in English.''} For \textbf{MMAU}, we use a fixed multiple-choice prompt that asks the model to return exactly one option letter in parentheses. The maximum number of generated tokens is 256 for captioning and 16 for MMAU. For the Audio Flamingo 3 validation, we use the official chat-template inference path and evaluate keep ratios 0.5 and 0.25, corresponding to moderate and aggressive compression, with the same compression interface where applicable.

Unless otherwise specified, the local window size is fixed to $w=8$ in the main experiments. We test keep ratios of 0.75, 0.5, and 0.25. AudioCaps~\citep{audiocaps} is evaluated on 883 unique test clips after deduplication, Clotho~\citep{clotho} on 1045 evaluation clips, and MMAU~\citep{mmau} on 1000 test-mini samples; these benchmarks complement broader audio-language evaluation efforts such as AudioBench~\citep{audiobench}. We report CIDEr on AudioCaps and Clotho, and overall accuracy on MMAU. Timing is measured on a single RTX 3090 with batch size 4.

\subsection{Compression Budget}

All methods are compared under the same keep ratio $\rho$, which controls the target number of active encoder-space audio tokens after compression. Given an original encoder-space sequence of length $L$, each method targets an active compressed length
\begin{equation}
L'=\max(1,\mathrm{round}(\rho L)).
\end{equation}
Thus, keep ratios of 0.75, 0.5, and 0.25 correspond to retaining approximately 75\%, 50\%, and 25\% of the original audio-token budget, respectively. Equivalently, they reduce the active audio-token budget by approximately 25\%, 50\%, and 75\%.

This convention provides a matched compression budget across merging, pruning, and pooling baselines. For LTBM and Global Merge, the budget controls the number of tokens after iterative bipartite merging. For Global Top-K and Segmentwise Top-K, it controls the number of retained tokens. For UniAvg, the budget is approximated by a discrete pooling factor; therefore, UniAvg is unavailable at keep ratio 0.75 in our implementation.

\subsection{Main Results on Captioning and MMAU}
\begin{figure*}[t]
    \centering
    \includegraphics[width=0.92\textwidth]{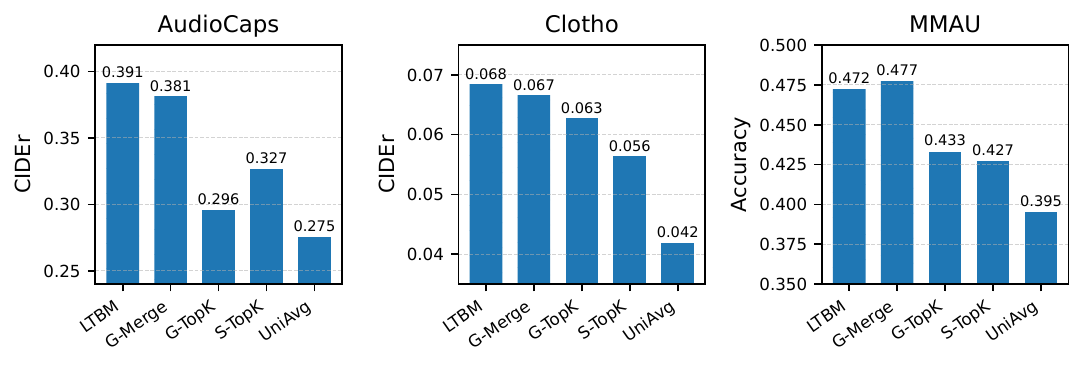}
    \caption{Performance under aggressive audio-token compression at keep ratio 0.25. Hatched bars denote the two bipartite merging variants. LTBM performs best on the two captioning benchmarks, while Global Merge performs best on MMAU. This pattern is consistent with the task-dependent locality effect observed in Tables~\ref{tab:caption_results} and~\ref{tab:mmau_results}.}
    \label{fig:aggressive_compression}
\end{figure*}
Table~\ref{tab:caption_results} reports the main captioning results. On \textbf{AudioCaps}, LTBM is strongest among the tested compression methods at keep ratios 0.75 and 0.25, while Global Merge is slightly stronger at 0.5. The most important pattern appears under aggressive compression: at keep ratio 0.25, LTBM reaches 0.3909 CIDEr, outperforming Global Merge (0.3809), Global Top-K (0.2957), Segmentwise Top-K (0.3267), and UniAvg (0.2754). On \textbf{Clotho}, compression is substantially more challenging for all methods, reflecting a harder captioning regime with finer-grained semantic requirements. We therefore interpret Clotho as a harder captioning stress test for comparing relative degradation patterns among compression methods. Even in this setting, LTBM remains best at keep ratios 0.75 and 0.25, while Global Merge is marginally stronger at 0.5. Across both captioning benchmarks, locality-aware merging is most competitive at keep ratios 0.75 and 0.25, suggesting that local temporal structure is a useful bias for preserving captioning quality under compression.

Table~\ref{tab:mmau_results} reports results on \textbf{MMAU}. Here the pattern differs from captioning: Global Merge is strongest at all keep ratios, and LTBM does not consistently exceed it. This contrast is itself informative. It suggests that the value of locality depends on the downstream task. Captioning benefits from preserving short-range acoustic continuity for open-ended language generation, whereas multiple-choice audio understanding is comparatively more compatible with globally aggregated evidence for closed-form decision making.

\begin{table*}[t]
\centering
\caption{Captioning results on AudioCaps and Clotho. We report CIDEr at keep ratios 0.75 / 0.5 / 0.25.}
\label{tab:caption_results}
\small
\begin{tabular}{lcccccc}
\toprule
\multirow{2}{*}{Method}
& \multicolumn{3}{c}{AudioCaps}
& \multicolumn{3}{c}{Clotho} \\
\cmidrule(lr){2-4}\cmidrule(lr){5-7}
& 0.75 & 0.5 & 0.25 & 0.75 & 0.5 & 0.25 \\
\midrule
Baseline & 0.4825 & 0.4825 & 0.4825 & 0.2830 & 0.2830 & 0.2830 \\
LTBM (local) & \textbf{0.4951} & 0.4782 & \textbf{0.3909} & \textbf{0.1838} & 0.1096 & \textbf{0.0683} \\
Global Merge & 0.4823 & \textbf{0.4805} & 0.3809 & 0.1812 & \textbf{0.1138} & 0.0666 \\
Global Top-K & 0.4912 & 0.4578 & 0.2957 & 0.1594 & 0.0906 & 0.0627 \\
Segmentwise Top-K & 0.4774 & 0.4728 & 0.3267 & 0.1639 & 0.0935 & 0.0564 \\
UniAvg & -- & 0.4748 & 0.2754 & -- & 0.0990 & 0.0419 \\
\bottomrule
\end{tabular}
\end{table*}

\begin{table}[t]
\centering
\caption{MMAU test-mini results. We report overall accuracy at keep ratios 0.75 / 0.5 / 0.25.}
\label{tab:mmau_results}
\small
\begin{tabular}{lccc}
\toprule
Method & 0.75 & 0.5 & 0.25 \\
\midrule
Baseline & 0.554 & 0.554 & 0.554 \\
LTBM (local) & 0.556 & 0.523 & 0.472 \\
Global Merge & \textbf{0.564} & \textbf{0.530} & \textbf{0.477} \\
Global Top-K & 0.493 & 0.466 & 0.433 \\
Segmentwise Top-K & 0.510 & 0.472 & 0.427 \\
UniAvg & -- & 0.502 & 0.395 \\
\bottomrule
\end{tabular}
\end{table}

\subsection{Task-Dependent Locality Effect}

The local-versus-global comparison is central to this work. Figure~\ref{fig:aggressive_compression} visualizes the aggressive-compression setting at keep ratio 0.25, where the difference between compression strategies is most visible. If the gains of LTBM came mainly from token merging rather than locality, then local and global merge would be expected to show more similar behavior. Table~\ref{tab:local_global} shows a more informative picture. On \textbf{AudioCaps} and \textbf{Clotho}, local merging is better at keep ratios 0.75 and 0.25, with the clearest benefit under stronger compression. On \textbf{MMAU}, by contrast, Global Merge is consistently stronger. This suggests that the effect is not simply ``merging helps,'' but that the usefulness of locality may depend on task structure.

A plausible explanation is that captioning and multiple-choice audio understanding emphasize different forms of compressed evidence. Audio captioning requires the model to preserve short-range acoustic continuity so that temporally extended events, such as sirens, rain, engines, or applause, remain semantically coherent after compression. In this regime, locality-aware merging is advantageous because it aggregates nearby similar tokens while preserving local event structure. By contrast, multiple-choice audio understanding often reduces to selecting the most discriminative evidence for a closed set of answers. In such cases, globally matched aggregation can be more effective because it is free to combine salient evidence across wider temporal ranges. Under this view, the difference between LTBM and Global Merge is not merely quantitative but reflects a task-dependent trade-off between local continuity preservation and global evidence aggregation.

The same perspective also helps interpret the gap between AudioCaps and Clotho. Although both are captioning benchmarks, Clotho is empirically a harder compression regime, likely because its captions often require finer-grained scene description and more detailed semantic composition. This makes all compression methods more sensitive on Clotho; nevertheless, LTBM remains relatively stronger among the tested methods at keep ratios 0.75 and 0.25.

\begin{table}[t]
\centering
\caption{LTBM minus Global Merge. Positive values favor local merge.}
\label{tab:local_global}
\small
\begin{tabular}{lccc}
\toprule
Dataset & 0.75 & 0.5 & 0.25 \\
\midrule
AudioCaps & +0.0128 & -0.0023 & +0.0100 \\
Clotho & +0.0026 & -0.0042 & +0.0017 \\
MMAU & -0.0080 & -0.0070 & -0.0050 \\
\bottomrule
\end{tabular}
\end{table}

\subsection{Cross-Backbone Validation on Audio Flamingo 3}

To examine whether the captioning-side behavior is specific to Qwen2-Audio, we further evaluate Audio Flamingo 3 as a second backbone. We use the same training-free compression interface where applicable: compression is applied in encoder space before the multimodal projector, all backbone parameters remain frozen, and the evaluation focuses on moderate and aggressive compression settings.

Table~\ref{tab:af3_sanity} reports results on AudioCaps and MMAU. On AudioCaps, LTBM outperforms Global Merge at both keep ratios. The margin is especially clear under aggressive compression: at keep ratio 0.25, LTBM obtains 0.2515 CIDEr compared with 0.2196 for Global Merge. LTBM is also the strongest method among the compared methods at keep ratios 0.5 and 0.25. These results provide additional evidence that the captioning-side advantage of locality-aware merging is not merely a Qwen2-Audio-specific artifact.

On MMAU, the local--global separation is less pronounced on Audio Flamingo 3. LTBM is slightly better at keep ratio 0.5, while Global Merge is slightly better at 0.25. This suggests that the local--global distinction on MMAU can be more backbone-dependent than the captioning-side behavior. We therefore treat the Audio Flamingo 3 results as a cross-backbone validation rather than as a complete second-backbone replication.

\begin{table}[t]
\centering
\caption{Cross-backbone validation on Audio Flamingo 3 at keep ratios 0.5 and 0.25. AudioCaps reports CIDEr and MMAU reports overall accuracy.}
\label{tab:af3_sanity}
\small
\setlength{\tabcolsep}{3.8pt}
\begin{tabular}{llcc}
\toprule
Dataset & Method & 0.5 & 0.25 \\
\midrule
\multirow{5}{*}{AudioCaps}
& Baseline & 0.2175 & 0.2175 \\
& LTBM & \textbf{0.3012} & \textbf{0.2515} \\
& Global Merge & 0.2912 & 0.2196 \\
& Global Top-K & 0.2736 & 0.1922 \\
& Segmentwise Top-K & 0.2716 & 0.2102 \\
\midrule
\multirow{5}{*}{MMAU}
& Baseline & 0.744 & 0.744 \\
& LTBM & \textbf{0.744} & 0.695 \\
& Global Merge & 0.736 & \textbf{0.699} \\
& Global Top-K & 0.698 & 0.675 \\
& Segmentwise Top-K & 0.713 & 0.673 \\
\bottomrule
\end{tabular}
\end{table}

\subsection{Case Study}

Table~\ref{tab:case_study} provides representative AudioCaps examples under aggressive compression at keep ratio 0.25. The examples are selected to compare what acoustic evidence is preserved by pruning, global merging, and locality-aware merging.

In the first example, LTBM preserves both the baby-crying event and the nearby human soothing context, while Global Merge retains the dominant cry but drops the surrounding speech context, and Global Top-K shifts toward a generic speech description. In the second example, all compressed methods struggle with the full scene, which contains engine noise, wind, water sounds, and overlapping speech. This illustrates a remaining failure mode: when several weak or overlapping sources are present, aggressive compression can collapse the scene into a partial speech-centric description. In the third example, the merging methods better preserve the goat event than Global Top-K, while LTBM produces a shorter but correct description of the dominant acoustic source. Overall, these cases suggest that locality-aware merging is helpful when the target event is embedded in short-range acoustic context, but it does not fully solve complex multi-source scene compression.

\begin{table*}[!tbp]
\centering
\caption{AudioCaps case study at keep ratio 0.25. We compare representative outputs from Global Top-K, Global Merge, and LTBM. S1--S3 correspond to AudioCaps samples ztSjcZNUY7A, D9tinq3RMpU, and JsoBpL86R5U.}
\label{tab:case_study}
\small
\setlength{\tabcolsep}{4pt}
\renewcommand{\arraystretch}{1.12}
\begin{tabular}{
>{\raggedright\arraybackslash}p{0.05\textwidth}
>{\raggedright\arraybackslash}p{0.25\textwidth}
>{\raggedright\arraybackslash}p{0.20\textwidth}
>{\raggedright\arraybackslash}p{0.20\textwidth}
>{\raggedright\arraybackslash}p{0.20\textwidth}
}
\toprule
Ex. & Reference summary & Global Top-K & Global Merge & LTBM \\
\midrule
S1 &
baby crying with a woman speaking or soothing in the background &
A woman is speaking to a baby. &
A baby crying. &
A baby crying and being consoled by a woman. \\
\midrule
S2 &
motorboat or ship engine with wind, water sounds, crowd speech, and distant male speech &
A man is speaking. &
A woman is speaking in a crowded indoor location. &
A man is speaking in the background. \\
\midrule
S3 &
people speaking with a goat bleating and child or background speech &
She's asking for help. &
A goat bleats as a woman speaks in the background. &
A goat bleats nearby. \\
\bottomrule
\end{tabular}
\end{table*}

\subsection{Window Ablation and Timing}

Table~\ref{tab:ablation_timing} reports the effect of varying the locality window on AudioCaps together with timing results. The $w=8$ column corresponds to the default setting used in the main experiments, while $w=4$ and $w=16$ provide additional matched comparisons around that default. Overall, small and moderate windows perform favorably, supporting the role of temporal locality. We therefore adopt $w=8$ as the default configuration in the main experiments because it delivers strong performance across compression ratios while maintaining a stable operating point.

Timing is measured after model loading and one warm-up batch on a 100-sample subset of AudioCaps at keep ratio 0.5. The reported value is the end-to-end per-sample runtime of our evaluation script, including preprocessing and generation with 32 new tokens. LTBM gives a modest runtime reduction from 0.2841 to 0.2592 seconds per sample. The speedup is not expected to scale one-to-one with the nominal token reduction because the evaluated integration uses a compatibility-preserving multimodal interface rather than modifying the language-model cache or attention kernels. Although UniAvg is faster, LTBM better preserves captioning quality under stronger compression.

\paragraph{Interpretation.}
The ablation suggests that LTBM is not highly sensitive to a single tuned window size. Small and moderate windows are consistently competitive, while $w=16$ is slightly weaker at keep ratios 0.75 and 0.5, suggesting that overly broad matching can dilute the locality constraint. At keep ratio 0.25, all windows perform similarly, indicating that the active token budget becomes the dominant factor under aggressive compression. We therefore use $w=8$ as a stable middle point rather than as a dataset-specific optimum.

The timing results should be interpreted under the compatibility-preserving inference path. LTBM reduces end-to-end runtime from 0.2841 to 0.2592 seconds per sample without modifying the language-model cache or attention kernels. Its main value is therefore a quality-efficiency trade-off: it reduces the active audio-token budget while preserving stronger captioning quality than fixed pooling or norm-based pruning under aggressive compression.

\begin{table}[!tbp]
\centering
\caption{Window ablation and end-to-end timing on AudioCaps.}
\label{tab:ablation_timing}
\small
\begin{tabular}{lccc}
\toprule
\multicolumn{4}{c}{\textbf{Window ablation (CIDEr)}} \\
\midrule
Keep ratio & $w=4$ & $w=8$ & $w=16$ \\
\midrule
0.75 & 0.4962 & 0.4951 & 0.4870 \\
0.50 & 0.4883 & 0.4782 & 0.4762 \\
0.25 & 0.3918 & 0.3909 & 0.3917 \\
\midrule
\multicolumn{4}{c}{\textbf{Timing at keep ratio 0.5 (sec/sample)}} \\
\midrule
Baseline & \multicolumn{3}{c}{0.2841} \\
LTBM & \multicolumn{3}{c}{0.2592} \\
Global Top-K & \multicolumn{3}{c}{0.2520} \\
UniAvg & \multicolumn{3}{c}{\textbf{0.2469}} \\
\bottomrule
\end{tabular}
\end{table}
\FloatBarrier
\section{Conclusion}

We presented \textbf{LTBM}, a training-free encoder-space audio-token compression method based on local temporal bipartite merging. Through a controlled comparison with Global Merge, we studied whether temporal locality is a useful inductive bias for audio-token compression. Experiments with Qwen2-Audio show a task-dependent locality effect: locality-aware merging better preserves captioning quality under stronger compression, while unconstrained global matching is more competitive for multiple-choice audio understanding. A cross-backbone validation on Audio Flamingo 3 further indicates that the captioning-side advantage of locality-aware merging persists under moderate and aggressive compression, although the local--global distinction on MMAU is less pronounced. These findings suggest that future audio-language inference systems should treat token locality not as a fixed assumption, but as a task-aware compression design choice.

\section*{Limitations}

This work focuses on training-free inference-time compression and does not update the audio encoder, multimodal projector, or language model. The current implementation uses a compatibility-preserving multimodal interface, so the measured latency improvement does not scale linearly with the nominal audio-token reduction. Our main experiments use Qwen2-Audio, with Audio Flamingo 3 used as a second-backbone validation under selected settings rather than a full replication across all datasets and compression ratios. 

\clearpage
\bibliography{refs}

\end{document}